\documentclass{llncs}
\usepackage{bm}
\usepackage{makeidx}  
\usepackage{amsmath}
\usepackage{amssymb}
\usepackage{hyperref}
\usepackage{graphicx}
\usepackage[T1]{fontenc}
\usepackage[noend]{algpseudocode}
\usepackage{algorithm}
\usepackage{algorithmicx}
\usepackage{subcaption}
\usepackage{paralist}
\usepackage{float}

\makeatletter
\renewcommand\paragraph{\@startsection{paragraph}{4}{\z@}%
                       {-2\p@ \@plus -4\p@ \@minus -4\p@}%
                       {-0.5em \@plus -0.22em \@minus -0.1em}%
                       {\normalfont\normalsize\bfseries}}
\renewcommand\section{\@startsection{section}{1}{\z@}%
                       {-12\p@ \@plus -4\p@ \@minus -4\p@}%
                       {8\p@ \@plus 4\p@ \@minus 4\p@}%
                       {\normalfont\large\bfseries\boldmath
                        \rightskip=\z@ \@plus 8em\pretolerance=10000 }}
\makeatother

\begin{document}
\frontmatter          
\pagestyle{headings}  
\mainmatter           

\setlength{\intextsep}{1\baselineskip}
\setlength{\floatsep}{1\baselineskip}
\setlength{\textfloatsep}{1\baselineskip}

\title{\textsc{ATPboost}:  Learning Premise Selection \\
in Binary Setting with ATP Feedback}
\titlerunning{ATPboost}  
\author{Bartosz Piotrowski\inst{1,2} \and Josef Urban\inst{1}\thanks{Supported by the \textit{AI4REASON} ERC Consolidator grant number 649043, and by the Czech project AI\&Reasoning CZ.02.1.01/0.0/0.0/15\_003/0000466 and the European Regional Development Fund.}}

\authorrunning{Bartosz Piotrowski \and Josef Urban} 
\institute{Czech Institute of Informatics, Robotics and Cybernetics, Prague,
Czech Republic
\and
Faculty of Mathematics, Informatics and Mechanics, University of Warsaw, Poland}

\maketitle            

\begin{abstract}
\textsc{ATPboost} is a system for solving sets of large-theory
problems by interleaving ATP runs with state-of-the-art machine
learning of premise selection from the proofs.  Unlike many previous
approaches that use multi-label setting, the learning is implemented
as binary classification that estimates the pairwise-relevance of
(\textit{theorem, premise}) pairs. 
\textsc{ATPboost} uses for this the XGBoost gradient boosting
algorithm, which is fast and has state-of-the-art performance on many
tasks. Learning in the binary setting however requires negative
examples, which is nontrivial due to many alternative proofs.  We
discuss and implement several solutions in the context of the ATP/ML
feedback loop, and show that \textsc{ATPboost} with such methods
significantly outperforms the k-nearest neighbors multilabel
classifier.

\end{abstract}
\section{Introduction: Machine Learning for Premise Selection}
\label{Intro}
Assume that $c$ is a conjecture which is a logical consequence of a large set of premises $P$.
The chance of finding a proof of $c$ by an automated theorem prover (ATP) often depends on choosing a small subset of $P$ relevant for proving $c$.
This is known as the \textit{premise selection} task~\cite{abs-1108-3446}.
This task is crucial to make ATPs usable for proof automation over large formal
corpora created with systems such as Mizar, Isabelle, HOL, and
Coq~\cite{hammers4qed}. Good methods for premise selection typically also
transfer to related tasks, such as \emph{internal proof guidance} of
ATPs~\cite{JakubuvU17a,KaliszykU15,LoosISK17,UrbanVS11} and \emph{tactical
guidance} of ITPs~\cite{GauthierKU17}.

The most efficient premise selection methods use \textit{data-driven/machine-learning} approaches.
Such methods work as follows. Let $T$ be a set of theorems with their proofs. Let $C$ be a set of conjectures without proofs, each associated with a set of available premises that can be used to prove them.
We want to learn a (statistical) model from $T$, which for each conjecture $c \in C$ will rank its available premises according  to their relevance for producing an ATP proof of $c$.
Two different machine learning settings can be used for this task:
\begin{compactenum}
        \item \emph{multilabel classification}: we treat premises used in the proofs as opaque labels and we create a model capable of labeling conjectures based on their features,
	\item \emph{binary classification}: here the aim of the learning model is to recognize pairwise-relevance of the (\textit{conjecture, premise}) pairs, i.e.\ to decide what is the chance of a premise being relevant for proving the conjecture based on the features of both the conjecture and the premise.
\end{compactenum}

Most of the machine learning methods for premise selection have so far
used the first
setting~\cite{BlanchetteGKKU16,holyhammer,mizar40}.  This includes
fast and robust machine learning algorithms such as \textit{naive
  Bayes} and \textit{K nearest neighbors} (k-NN) capable of multilabel
classification with many examples and labels. This is needed for large
formal libraries with many facts and proofs.
There are however several reasons why the second approach may be better:
\begin{compactenum}
\item Generality: in binary classification it is easier to estimate
the relevance of (\textit{conjecture, premise}) pairs where the premise
  was so far unseen (i.e., not in the training data).
        \item State-of-the-art ML algorithms are often capable of
          learning subtle aspects of complicated problems based on the
          features. The multilabel approach trades the rich feature
          representation of the premise for its opaque label.
        \item Many state-of-the-art ML algorithms are binary
          classifiers or they struggle when performing multilabel
          classification for a large number of labels.
\end{compactenum}
Recently, substantial work~\cite{deepmath} has been done in the
binary 
setting.  In particular, applying deep learning to premise selection
has improved state of the art in the field. There are however
modern and efficient learning algorithms such as XGBoost~\cite{xgboost} that are much
less computationally-intensive then deep learning methods. Also,
obtaining negative examples for training the binary classifiers is a
very interesting problem in the context of many alternative ATP proofs
and a feedback loop between the ATP and the learning system.


\subsection{Premise Selection in Binary Setting with Multiple Proofs}
The existence of multiple ATP proofs makes premise selection different from conventional machine learning applications. This is evident 
especially in the binary classification setting.
The ML algorithms for recognizing pairwise relevance of (\textit{conjecture, premise}) pairs require good data consisting of two (typically balanced) classes of positive and negative examples. 
But there is no conventional way how to construct such data in our domain.
For every true conjecture there are infinitely many mathematical proofs. The ATP proofs are often based on many different sets of premises. The notions of \textit{useful} or
\textit{superfluous premise} are only approximations of their counterparts defined for sets of premises.

As an example, consider the following frequent situation: 
a conjecture $c$ can be ATP-proved with two sets of axioms: $\{p_1, p_2\}$ and $\{p_3, p_4, p_5\}$.
Learning only from one of the sets as positives and presenting the other as negative (\textit{conjecture, premise}) pairs 
may considerably distort the learned notion of a \emph{useful premise}.
This differs from the multilabel setting, where negative data are typically not used by the fast ML algorithms such as naive Bayes and k-NN. They just aggregate different positive examples into the final ranking.

Therefore, to further improve the premise selection algorithms it seems useful to
consider learning from multiple proofs and to develop methods producing good negative data.
The most suitable way how to do that is to allow multiple interactions of the machine learner with the ATP system.
In the following section we present the \textsc{ATPboost} system, which implements several such algorithms.


\section{\textsc{ATPboost:} Setting, Algorithms and Components}
\textsc{ATPboost}\footnote{The Python package is at
  \url{https://github.com/BartoszPiotrowski/ATPboost}.} is a system for
solving sets of large-theory problems by interleaving ATP runs with
learning of premise selection from the proofs using the state-of-the-art XGBoost algorithm. The system
implements several algorithms and consists of several components
described in the following sections.
%
%
Its setting is a large theory $\mathcal{T}$, extracted from a large ITP library where facts appear in a chronological order.
In more detail, we assume 
the following inputs and notation:
\begin{compactenum}
	\item $T$ -- names of theorems (and problems) in a large theory $\mathcal{T}$.
	\item $P$ -- names of all facts (premises) in $\mathcal{T}$. We require $P \supseteq T$.
	\item \textsc{Statements}$_P$ of all $p \in P$ in the TPTP format~\cite{Sutcliffe09a} .
	\item \textsc{Features}$_P$ -- characterizing each $p \in P$. Here we use the same features as in~\cite{mizar40} and write
$\bm{f}_p$ for the (sparse) vector of features of $p$.
	\item \textsc{Order}$_P$ ($<_P$) -- total order on $P$; $p$ may be used to prove $t$ iff $p <_P t$. We write $A_t$ for $\{p: p <_P t\}$, i.e. the set of premises allowed for $t$.
	\item \textsc{Proofs}$_{T'}$ for a subset $T' \subseteq T$. Each $t \in T'$ may have many proofs -- denoted by $\mathcal{P}_t$. $P_t$ denotes the premises needed for at least one proof in $\mathcal{P}_t$.
        \end{compactenum}

\subsection{Algorithms}
We first give a high-level overview and pseudocode of the algorithms implemented in
\textsc{ATPboost}. Section~\ref{par:proc} then describes the used components in detail.
\begin{compactenum}
\item[\textbf{Algorithm~\ref{alg:split}}] is the simplest setting.
Problems are split into the train/test sets,
XGBoost learns from
the training proofs, and its predictions are ATP-evaluated on the test set.
This is used mainly for parameter optimization.
\item[\textbf{Algorithm \ref{alg:adding}}] evaluates the trained XGBoost also on the training
part, possibly finding new proofs that are used to update the training
data for the next iteration. The test problems and proofs are
never used for training. Negative mining may be used to find the worst misclassified premises and to correspondingly update
the training data in the next iteration.

\item[\textbf{Algorithm \ref{alg:loop}}] begins with no training set, starting with ATP runs on random rankings.
XGBoost is trained on the ATP proofs from the previous iteration, producing new ranking for all problems
for the next iteration. This is a MaLARea-style~\cite{US+08} feedback loop between the ATP and the learner.
\end{compactenum}

\subsection{Components}
\label{par:proc}
Below we describe the main components of the \textsc{ATPboost} algorithms and the main ideas behind them. As discussed in Section~\ref{Intro}, they take into account the binary learning setting, and in particular implement the need to teach the system about multiple proofs by proper choice of examples, continuous interaction with the ATP and intelligent processing of its feedback.
The components are available as procedures in our Python package.




\begin{algorithm}
\scriptsize
\caption{\small{Simple training/test split.}}
\label{alg:split}
\begin{algorithmic}[1]
	\Require Set of theorems $T$, set of premises $P \supseteq T$,
			\textsc{Proofs}$_T$,
			\textsc{Features}$_P$,
			\textsc{Statements}$_P$,
			\textsc{Order}$_P$,
			\textsc{params}$_\text{set}$,
			\textsc{params}$_\text{model}$.
	\State $T_\text{train}, T_\text{test} \gets \textsc{RandomlySplit}(T)$
	\State $\mathcal{D} \gets
		\textsc{CreateTrainingSet}(
						\textsc{Proofs}_{T_\text{train}},
						\textsc{Features}_P,
						\textsc{Order}_P,
						\textsc{params}_\text{set})$
	\State $\mathcal{M} \gets \textsc{TrainModel}(
						\mathcal{D},
						\textsc{params}_\text{model})$
	\State $\mathcal{R} \gets \textsc{CreateRankings}(
						T_\text{test},
						\mathcal{M},
						\textsc{Features}_P,
						\textsc{Order}_P)$
	\State $\mathcal{P} \gets \textsc{ATPevaluation}(
						\mathcal{R},
						\textsc{Statements}_P)$
\end{algorithmic}
\end{algorithm}
\begin{algorithm}
\scriptsize
\caption{\small{Incremental feedback-loop with training/test split.}}
\label{alg:adding}
\begin{algorithmic}[1]
	\Require Set of theorems $T$, set of premises $P \supseteq T$,
			\textsc{Features}$_P$,
			\textsc{Statements}$_P$,
			\textsc{Proofs}$_T$,
			\textsc{Order}$_P$,
			\textsc{params}$_\text{set}$,
			\textsc{params}$_\text{model}$,
			\textsc{params}$_\text{negmin}$ (optionally).
	\State $T_\text{train}, T_\text{test} \gets \textsc{RandomlySplit}(T)$
	\State $\mathcal{D} \gets
		\textsc{CreateTrainingSet}(
					\textsc{Proofs}_{T_\text{train}},
					\textsc{Features}_P,
					\textsc{Order}_P,
					\textsc{params}_\text{set})$
	\Repeat
		\State $\mathcal{M} \gets \textsc{TrainModel}(
						\mathcal{D},
						\textsc{params}_\text{model})$
		\State $\mathcal{R}_\text{train} \gets \textsc{CreateRankings}(
						T_\text{train},
						\mathcal{M},
						\textsc{Features}_P,
						\textsc{Order}_P)$
		\State $\mathcal{R}_\text{test} \gets \textsc{CreateRankings}(
						T_\text{test},
						\mathcal{M},
						\textsc{Features}_P,
						\textsc{Order}_P)$
		\State $\mathcal{P}_\text{train} \gets \textsc{ATPevaluation}(
						\mathcal{R}_\text{train},
						\textsc{Statements}_P)$
		\State $\mathcal{P}_\text{test} \gets \textsc{ATPevaluation}(
						\mathcal{R}_\text{test},
						\textsc{Statements}_P)$
		\State $\textsc{Update}(\textsc{Proofs}_\text{train},
						\mathcal{P}_\text{train})$
		\State $\textsc{Update}(\textsc{Proofs}_\text{test},
						\mathcal{P}_\text{test})$
		\If{\textsc{params}$_\text{negmin}$}
			\State $\mathcal{D} \gets
				\textsc{NegativeMining}(
						\mathcal{R},
						\textsc{Proofs}_{\text{train}},
						\textsc{Features}_P,
						\textsc{Order}_P,
						\textsc{params}_\text{negmin})$
		\Else
			\State $\mathcal{D} \gets
				\textsc{CreateTrainingSet}(
						\textsc{Proofs}_{\text{train}},
						\textsc{Features}_P,
						\textsc{Order}_P,
						\textsc{params}_\text{set})$
		\EndIf
	\Until{Number of \textsc{Proofs}$_\text{test}$ increased after
						\textsc{Update}.}
\end{algorithmic}
\end{algorithm}
\begin{algorithm}[htbp!]
\scriptsize
\caption{\small{Incremental feedback-loop starting with no proofs.}}
\label{alg:loop}
\begin{algorithmic}[1]
	\Require Set of theorems $T$, set of premises $P \supseteq T$,
			\textsc{Features}$_P$,
			\textsc{Statements}$_P$,
			\textsc{Order}$_P$,
			\textsc{params}$_\text{set}$,
			\textsc{params}$_\text{model}$,
			\textsc{params}$_\text{negmin}$ (optionally).
	\State $\textsc{Proofs}_T \gets \emptyset$
	\State $\mathcal{R} \gets \textsc{CreateRandomRankings}(T)$
	\State $\mathcal{P} \gets \textsc{ATPevaluation}(
						\mathcal{R},
						\textsc{Statements}_P)$
	\State $\textsc{Update}(\textsc{Proofs}_T, \mathcal{P})$
	\State $\mathcal{D} \gets
		\textsc{CreateTrainingSet}(
					\textsc{Proofs}_{T},
					\textsc{Features}_P,
					\textsc{Order}_P,
					\textsc{params}_\text{set})$
	\Repeat
		\State $\mathcal{M} \gets \textsc{TrainModel}(
						\mathcal{D},
						\textsc{params}_\text{model})$
		\State $\mathcal{R} \gets \textsc{CreateRankings}(
						T,
						\mathcal{M},
						\textsc{Features}_P,
						\textsc{Order}_P)$
		\State $\mathcal{P} \gets \textsc{ATPevaluation}(
						\mathcal{R},
						\textsc{Statements}_P)$
		\State $\textsc{Update}(\textsc{Proofs}_T,
						\mathcal{P})$
		\If{\textsc{params}$_\text{negmin}$}
			\State $\mathcal{D} \gets
				\textsc{NegativeMining}(
						\mathcal{R},
						\textsc{Proofs}_{T},
						\textsc{Features}_P,
						\textsc{Order}_P,
						\textsc{params}_\text{negmin})$
		\Else
			\State $\mathcal{D} \gets
				\textsc{CreateTrainingSet}(
						\textsc{Proofs}_{T},
						\textsc{Features}_P,
						\textsc{Order}_P,
						\textsc{params}_\text{set})$
		\EndIf
	\Until{Number of \textsc{Proofs}$_T$ increased after \textsc{Update}.}
\end{algorithmic}
\end{algorithm}

\paragraph{\textsc{CreateTrainingSet}(\textsc{Proofs}$_T$, \textsc{Features}$_P$,
	\textsc{Order}$_P$, \textsc{params}).}
This procedure constructs a \textsc{TrainingSet} for a binary learning algorithm.
This is a sparse matrix of positive/negative examples and a corresponding vector of
binary labels.
The examples (matrix rows) are created
from \textsc{Proofs}$_{T}$ and 
\textsc{Features}$_P$, respecting \textsc{Order}$_{P}$.
Each example is a concatenation of $\bm{f}_t$ and $\bm{f}_p$, i.e., the features of a theorem $t$ and a premise $p$.
Positive examples express that $p$ is relevant for proving $t$, whereas the negatives mean the opposite.

The default method (\textsc{simple}) creates positives from all pairs $(t, p)$ where $p \in P_t$.
Another method (\textsc{short}) creates positives only from the \emph{short} proofs of $t$. These are the proofs
of $t$ with at most  $m+1$ premises, where $m$ is the minimal number of premises used in a proof from $\mathcal{P}_t$.
Negative examples for theorem $t$ are chosen randomly from pairs $(t, p)$ where
$p \in A_t \setminus P_t$.
The number of such
randomly chosen pairs is $\textsc{ratio} \cdot N_{\text{pos}}$, where
$N_{\text{pos}}$ is the number of positives and \textsc{ratio}$\in \mathbb{N}$ is
a parameter that needs to be optimized experimentally.
Since $|A_t \setminus
P_t|$ is usually much larger than $|P_t|$, it seems reasonable to have a large
\textsc{ratio}. This however increases class imbalance and the probability of
presenting to the learning algorithm a \textit{false negative}. This is
a pair $(t, p)$ where $p \notin P_t$, but there is an ATP proof of $t$ using $p$
that is not yet in our dataset.

\paragraph{\textsc{TrainModel}(\textsc{TrainingSet}, \textsc{params}).}

This procedure trains a binary learning classifier on the \textsc{TrainingSet}, creating a \textsc{Model}.
We use XGBoost~\cite{xgboost} -- a state-of-the-art tree-based gradient boosting
algorithm performing very well in machine learning competitions. It is also much faster to train compared to deep learning methods,
performs well with unbalanced training sets, and is optimized for working with sparse data.
XGBoost has several important parameters, such as
\textsc{numberOfTrees}, \textsc{maxDepth} (of trees) and \textsc{eta}
(learning rate). These parameters have significant influence on
the performance and require tuning. 

\paragraph{\textsc{CreateRankings}($C$, \textsc{Model},
	\textsc{Features}$_P$, \textsc{Order}$_P$).}

This procedure uses the trained \textsc{Model} to construct
      \textsc{Rankings}$_C$ of premises from $P$ for conjectures
      $c \in C \subseteq T$.
Each conjecture $c$ is paired with each premise $p <_P c$ and concatenations of $\bm{f}_c$ and $\bm{f}_p$
are passed to the \textsc{Model}. The \textsc{Model} outputs a real number in $[0, 1]$,
which is interpreted as the relevance of $p$ for proving $c$. The relevances are then used to sort the premises
into \textsc{Rankings}$_C$.

\paragraph{\textsc{ATPevaluation}(\textsc{Rankings}, \textsc{Statements}).}

Any ATP can be used for evaluation. By default we use E~\cite{Sch02-AICOMM} \footnote{The default time limit is 10 seconds and the memory limit is 2GB. The exact default command is: \texttt{./eprover --auto-schedule
--free-numbers -s -R --cpu-limit=10 --memory-limit=2000 --print-statistics -p
--tstp-format problem\_file}}.
As usual, we construct the ATP problems for several top slices (lengths $1, 2, \ldots, 512$) of the
\textsc{Rankings}.
To remove redundant premises we \textit{pseudo-minimize} the proofs: only the premises needed in the proofs are used as axioms and the ATP is rerun until a fixpoint is reached.

\paragraph{\textsc{Update}(\textsc{OldProofs}, \textsc{NewProofs}).}
The \textsc{Update} makes a union of the new and old proofs, followed by a subsumption reduction.
I.e., if premises of two proofs of $t$ are in a superset relation, the proof with the larger set is removed.


\paragraph{\textsc{NegativeMining}(\textsc{Proofs}$_T$, \textsc{Rankings}$_T$,
	\textsc{Features}$_P$, \textsc{Order}$_P$,\\ \textsc{params}).}


This is used as a more advanced alternative to \textsc{CreateTrainingSet}.
It examines the last
\textsc{Rankings}$_T$ for the most \textit{misclassified positives}. I.e., for each $t \in T$ we create a set $\mathit{MP}_t$
of those $p$ that were previously ranked high for $t$, but no ATP proof of $t$ was using $p$.
We define three variants:
\begin{compactenum}
	\item \textsc{negmin{\_}all}: Let $m_t$ be the maximum rank of a $t$-useful premise ($p \in P_t$) in
		\textsc{Rankings}$_T[t]$. Then $\mathit{MP}^1_t = \{p: rank_t(p)<m_t \land p \notin P_t  \}$.
	\item \textsc{negmin\_rand}: We randomly choose into $\mathit{MP}^2_t$ only a half of  $\mathit{MP}^1_t$.
        \item \textsc{negmin\_1}: $\mathit{MP}^3_t = \{p: rank_t(p)<|P_t| \land p \notin P_t  \}$.
              \end{compactenum}
The set $\mathit{MP}^i_t$ is then added as negatives to the examples produced by the \textsc{CreateTrainingSet} procedure.
The idea of such negative mining is that the learner takes into account the mistakes it made in the previous iteration.

\section{Evaluation}
We evaluate\footnote{All the scripts we used for the evaluation are available at
\url{https://github.com/BartoszPiotrowski/ATPboost/tree/master/experiments}}
the algorithms on a set of 1342
MPTP2078~\cite{abs-1108-3446} large (\emph{chainy}) problems that are
provable in 60s using their small (\emph{bushy})
versions.

\begin{figure}
	\centering
	\includegraphics[scale=0.55]{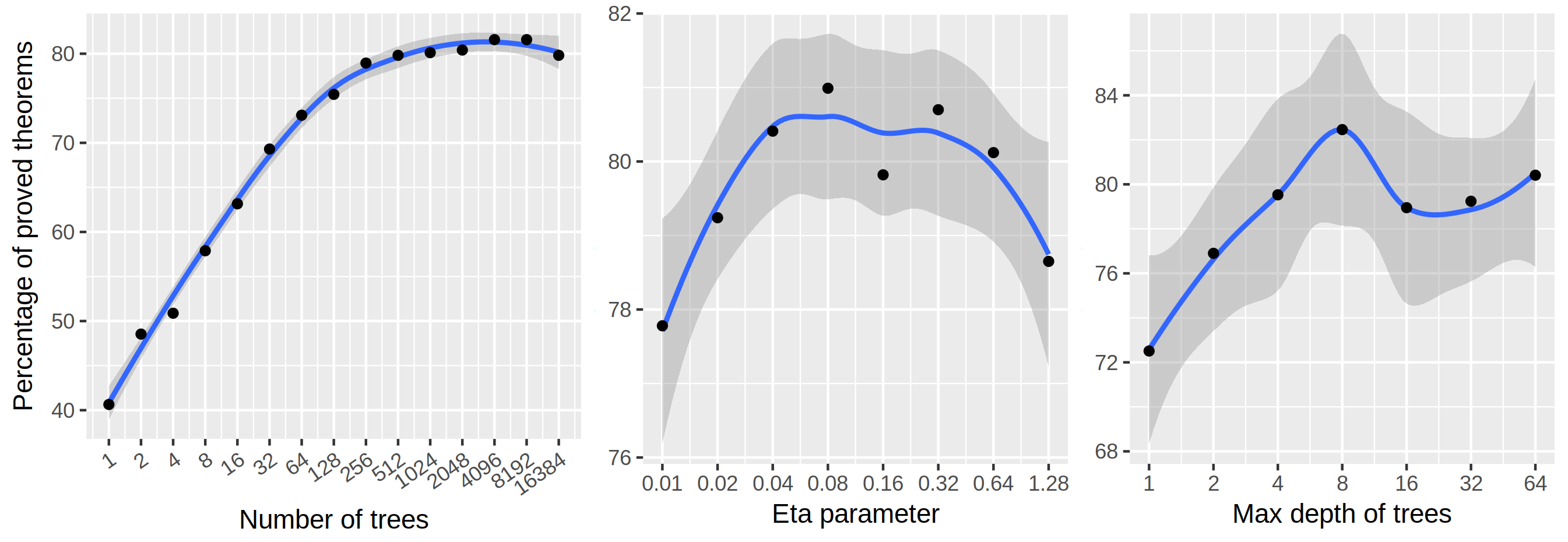} 
	\caption{\label{fig:grid} ATP performance of different
	parameters of the XGBoost model.}
\end{figure}

\paragraph{Parameter tuning:}
First we run Algorithm \ref{alg:split} to optimize the parameters.
The dataset was randomly split into a train  set of 1000 problems and test set of 342.
For the train set, we use the proofs obtained by the 60s run on the bushy versions.
We tune the \textsc{ratio} parameter of \textsc{CreateTrainingSet}, and the
\textsc{numberOfTrees}, \textsc{maxDepth} and \textsc{eta}  parameters of \textsc{TrainModel}.
Due to resource constraints we \textit{a priori} assume good defaults: \textsc{ratio} $=16$, \textsc{numberOfTrees} $=2000$,
\textsc{maxDepth} $=10$, \textsc{eta} $=0.2$. Then we observe how changing each parameter separately influences the results.
Table
\ref{tab:ratios} shows the ATP results for the \textsc{ratio} parameter, and Figure
\ref{fig:grid} for the model parameters.
\begin{table}[]
\centering
\begin{tabular}{l|ccccccc}
	\textsc{ratio}     & 1       & 2       & 4       & 8       & 16      & 32      & 64      \\ \hline
Proved (\%) & 74.0 & 78.4 & 79.0 & 78.7 & 80.1 & 79.8 & 80.1
\end{tabular}
\caption{\label{tab:ratios}Influence of the \textsc{ratio} of randomly generated negatives to positives.}
\end{table}
It is clear that a high number of negatives is important. Using \textsc{ratio} $=16$ proves
$6\%$ more test problems than the balanced setting (\textsc{ratio} $=1$).
It is also clear that a higher number of
trees -- at least 500 -- improves the results. However, too many trees (over 8000) slightly decrease the performance,
likely due to overfitting.
The \textsc{eta} parameter gives best results with values between $0.04$ and $0.64$, and
the \textsc{maxDepth} of trees should be around $10$.

We evaluate Algorithm~\ref{alg:split} also on a much bigger ATP-provable part of MML
with $29271$ theorems in train part and $3253$ in test. With parameters
\textsc{ratio} $=20$, \textsc{numberOfTrees} $=4000$, \textsc{maxDepth} $=10$
and \textsc{eta} $= 0.2$ we proved $58.78\%$ theorems (1912).  This is a 15.7\% improvement
over k-NN, which proved $50.81\%$ (1653) theorems. For a comparison, the improvement over k-NN obtained
(with much higher ATP time limits) with deep learning in~\cite{deepmath} was 4.3\%.


\paragraph{Incremental feedback loop with train/test split:}
This experiment evaluates Algorithm \ref{alg:adding}, testing different methods of negative
mining. 
The train/test split and the values of the parameters \textsc{ratio},
\textsc{numberOfTrees}, \textsc{maxDepth}, \textsc{eta} are taken from the
previous experiment.
We test six methods in parallel. Two \textsc{XGB} methods (\textsc{simple} and \textsc{short}) are the variants of the \textsc{CreateTrainingSet} procedure, three \textsc{XGB} methods (\textsc{negmin\_all}, \textsc{negmin\_rand} and \textsc{negmin\_1}) are the variants of \textsc{NegativeMining}, and the last one is a k-NN learner similar to the one from
		\cite{mizar40}, used here for comparison.

\begin{figure}
	\centering
	\includegraphics[width=0.7\textwidth]{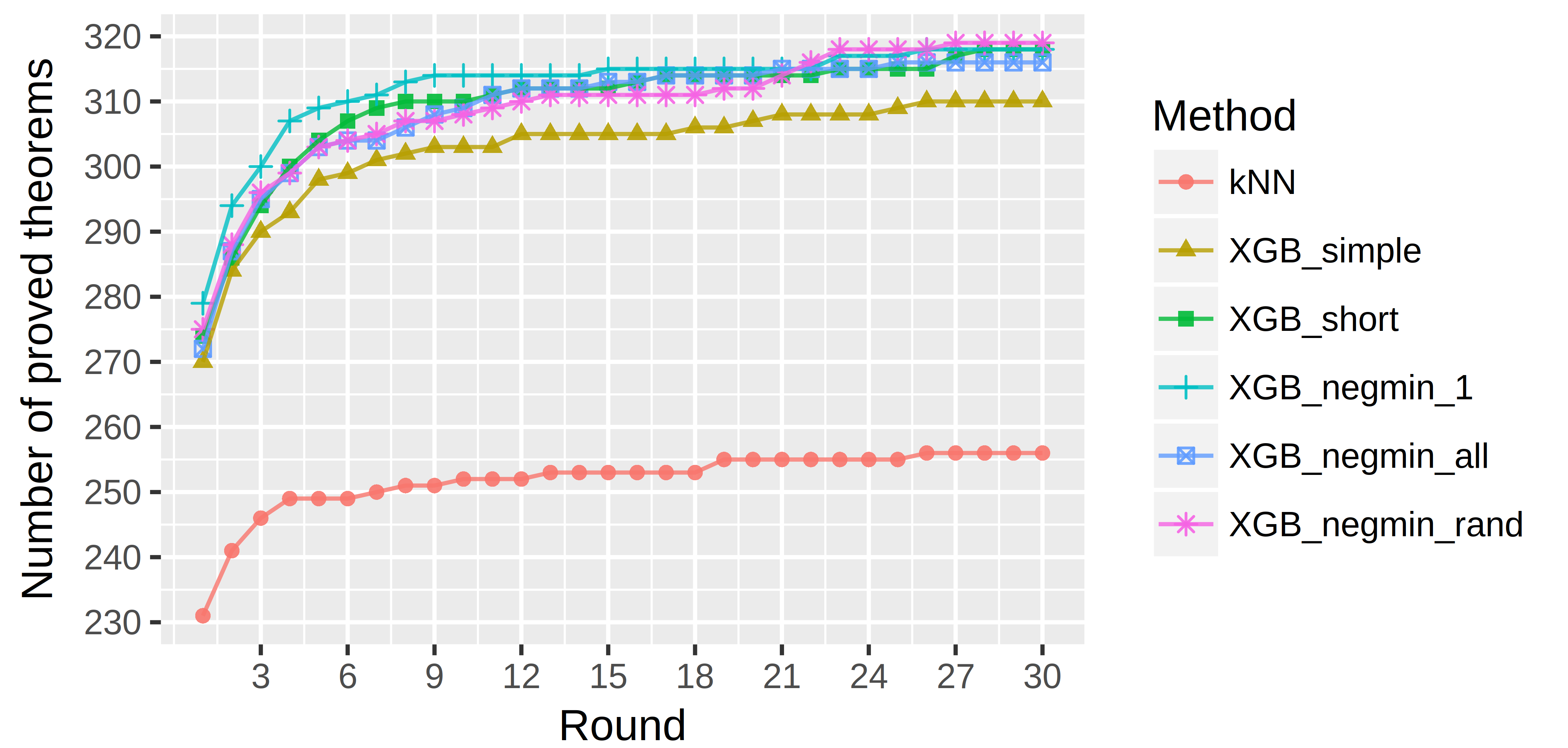} 
	\caption{\label{fig:adding} Number of proved theorems in subsequent
			iterations of Algorithm 2.}
\end{figure}

The experiment starts with the same proofs for training theorems as in the
previous one, and we performed 30 rounds of the feedback loop.
Figure \ref{fig:adding} shows the results.
All the new methods largely outperform k-NN, and
\textsc{XGB\_short} is much better than \textsc{XGB\_simple}.
I.e., positives from too many proofs seem harmful, as in \cite{KuhlweinU12b} where
this was observed with k-NN.
The differences between the \textsc{XGB} variants
\textsc{short}, \textsc{negmin\_1}, \textsc{negmin\_all},
and \textsc{negmin\_rand} do not seem significant
and all perform well. At the end of the loop (30th
round) 315-319 theorems of the 342 (ca $93\%$) are proved.

\begin{figure}
\centering
	\includegraphics[width=1.02\textwidth]{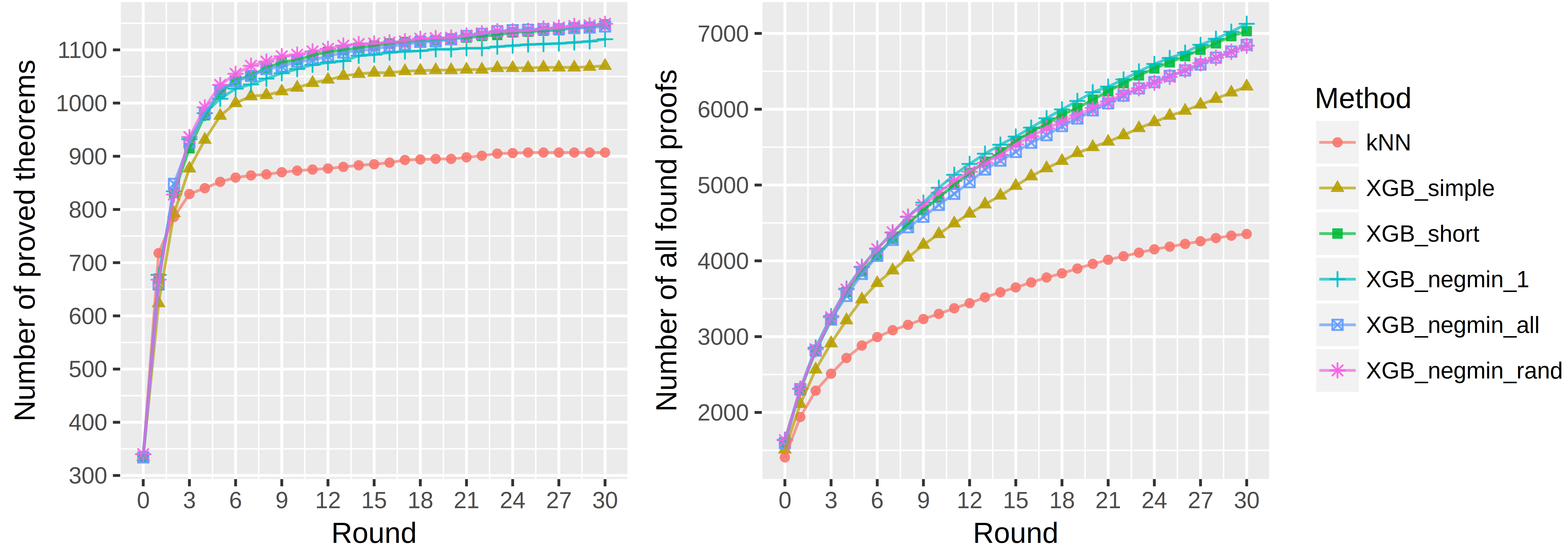}
	\caption{\label{fig:loops} Number of proved theorems (left) and number
	of all found proofs (right) in subsequent rounds of the experiment
	corresponding to Algorithm 3.}
\end{figure}

\paragraph{Incremental feedback-loop with no initial proofs:}
This is the final 
experiment which corresponds to the Algorithm
\ref{alg:loop} 
-- there is no train/test
split and no initial proofs. The first ATP evaluation
is done on random rankings, proving
335 simple theorems out of the 1342. Than
the feedback loop starts running with the same options as in the previous experiment.
Fig. \ref{fig:loops} shows the numbers of theorems that were proved
in the subsequent rounds, as well as the growth of the total number of different
proofs. This is important, because all these proofs are taken into account by the machine learning.
Again, k-NN is the weakest and \textsc{XGB\_simple} is worse than
the rest of the methods, which are statistically indistinguishable. In the last round \textsc{XGB\_negmin\_rand}
proves 1150 ($86\%$) theorems. This is 26.8\% more than k-NN (907) and 7.7\% more than \textsc{XGB\_simple} (1068).

\bibliographystyle{abbrv}
\bibliography{references,ate11}

\end{document}